# Early Detection of Alzheimer's - A Crucial Requirement

Brain is the central part of the human nervous system and any abnormality caused by any disease can lead to complete failure of human structural function. Alzheimer, an old age disease of people over 65 years [1] causes problems with memory, thinking and behavior. This disease progresses very slow and its identification in early stages is very difficult. It is not a specific disease and the patients may have problems with memory, communication, concentrated attention, reasoning, judgment, focusing, and visual perception. The symptoms appear slowly and these gradually will have worse effects. In its early stages, not only the patients themselves but their loved ones are generally unable to accept that the patient is suffering from disease.

Alzheimer's patients forget the recent information and face challenges in simple arithmetic. They also have problems in speaking and writing, misplace things and have difficulty in retracing. Their interest in job and social events lessens and their mood becomes unpredictable. One can observe visible changes in the personality of the patients. On average patients live of eight years after identification of Alzheimer, but patients survive from 4 to 20 years depending on their age and other health conditions [2].

In United States, Alzheimer's is ranked sixth disease in causing death and 5th for those patients who have 65 years of age and more [3]. During 2000 and 2008, 66 percent increase in deaths due to Alzheimer's has been seen, whereas the deaths from heart disease (ranked number one cause of death) have decreased [4]. Presently, in US number of Alzheimer's patients is over 5 million. One third senior citizens of US die with this disease. By 2050, the number of patients is expected 11 to 16 million, whereas a new patient in every 33 seconds and one million new patients are expected every year [4]. In 2013, expected cost caused to US is $203 billion which may rise to $1.2 trillion by 2050. Dr. Francis Collins, Director National Institutes of Health, US has designated $40 million from his fiscal year 2013 director's budget for Alzheimer's research. Collins also pointed out that President Barack Obama's FY14 budget would add $80 million for Alzheimer's research "over and above what's being supported" reflecting the way US administration prioritizes this disease.

In 1906, German psychiatrist and neuropathologist Alois Alzheimer described it first time in 1906 [5]. Exact cause of Alzheimer's is still to be known. Generally, it is caused due to loss of neurons and synapses in the cerebral cortex. Certain sub-cortical regions cause degeneration in the temporal lobe and parietal lobe, and parts of the frontal cortex [6]. Degeneration is also present in brainstem nuclei like the locus coeruleus [7]. MRI [Magnetic Resonance Images] and PET [Positron Emission Tomography] images are being used to determine the degeneration and reductions in the size of specific brain regions in people with Alzheimer's and comparing it with healthy people of their age [8]. Biomarkers on MRIs and images with PET provide reliable information in identifying the patients' stages of disease. Early detection of Alzheimer's and its stages is very important, because as it worsens it has no cure and patients have very dreadful life before their death.

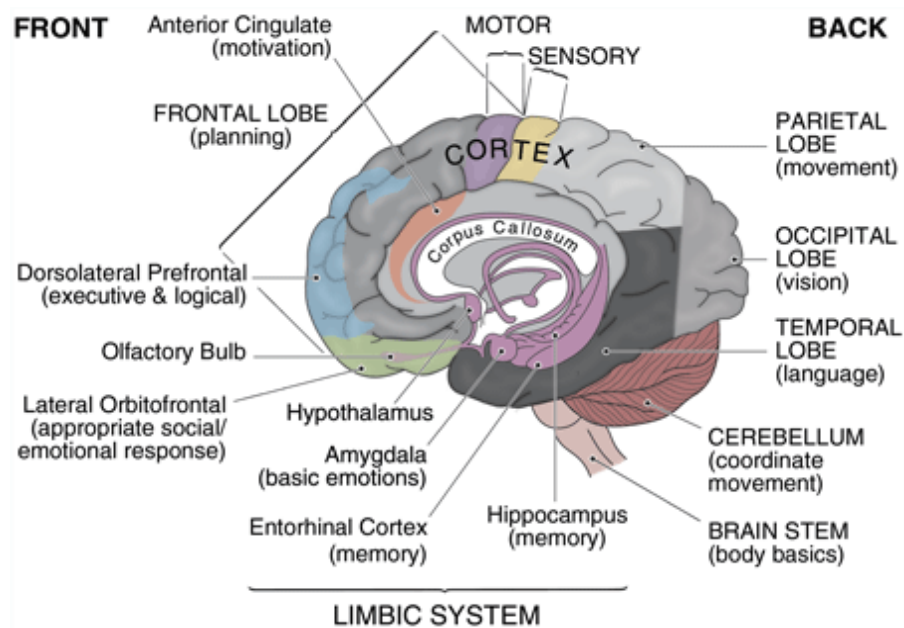

Courtesy: http://science.howstuffworks.com

Detection of Alzheimer's is very difficult in first four stages. In Stage1 and Stage 2, the patient does not experience any memory problems, whereas in Stage 3 mild cognitive decline can be diagnosed in some cases. In Stage 4, some visible changes like forgetting recent events and problems in management of financial events are seen in the patients. In Stage 5, the patients have moderately severe cognitive decline and need help in performing routine tasks. In Stage 5, the patients have severe cognitive decline and have awareness problems. In Stage 6, they forget their personal history. In Stage 7, the patients have serious problems. At this stage, they need greater help in performing routine tasks like eating or using the toilet. They cannot sit without support and have difficulty in holding their heads up. They develop abnormal reflexes and even have a difficulty to smile [2]. It is very important to note that Alzheimer's patients may face fatal outcomes along with non-fatal consequences like stroke, congestive heart failure, and ischemic attacks as a result of side effects of their treatment.

Detection of the Alzheimer's is done by clinical, neuropsychological, and neuro-imaging assessments. Atrophy calculation is an effective approach at late stages of the disease. The pathologic hallmarks of the disease are beta-amyloid (Aß) plaques, neurofibrillary tangles (NFTs), and reactive gliosis [9]. Researchers are making efforts for early detection of Alzheimer's, its cure or at least slow it down. Biomarkers in neuro-imaging provide an acceptable level of assistance in detection of disease and help in excluding other causes of dementia. Ashraf Awan et al [10] have proposed an image processing technique to identify brain abnormalities. They calculate the black and white pixels and compare these to identify the abnormalities.

Y. Liu et al [11] have presented an approach to discover biomarkers to predict Alzheimer's Dementia by using MRIs. This method is noninvasive and biomarkers are set to differentiate gray and white matter parenchyma and cerebrospinal fluid (CSF) filled spaces of the brain. It works in four steps: Image alignment using intensity normalization, midsagittal plane extraction, and affine transformation; feature extraction; feature screening; and subspace biomarker learning and cross validation using 40-fold leave ten out cross validation. The experiment was performed on a cluster containing 13 PCs, which had 2.4 GHz processors, 1 GB dual channel memory, and 80 GB hard drive that took 7 hours to register MRIs and 25 hours for feature generation. It gives encouraging results (over 90% accuracy) to set biomarkers for image classification.

AmirEhsan Lashkari [12] has used neural network, a feature selection method to classify the brain tissues. This method clearly describes normal and abnormal brain tissues using Gabor wavelets. This method saves the radiologist time, increases accuracy and yield of diagnosis. He has also reduced the feature space. MR images that have been used by him are T1_weighted, T2_weighted and PD images. T1_weighted MR Images shows the hard tissue darker gray scale density than neighbor tissues and in T2_weighted and PD, MR images the hard tissue is brighter gray scale density than neighbor tissues but in all MR images modalities, the normal tissues are almost similarity gray scale density.

The proposed method estimates defiant structures with the help of neural network and some statistical functions, in the first step. These structures are considered to be potential places that may have tumor. In second step, a combination of Gabor wavelets and neural network is used to confirm the presence of tumor or otherwise. Finally, tumor location and volume are computed. It is a better method as compared to traditional manual methods to detect the presence of abnormality in the human brain. It is a noninvasive procedure and in some cases it may predict false abnormality.

Ramaswamy Reddy et al [13] have proposed Spatial FCM (Fuzzy C-means) to classify the similar data points into clusters of feature space. The advantage of this method is that it allows pixels to be in multiple classes as compared to the traditional classification techniques which place each pixel in a specific class. In this algorithm, the probability of belonging of a pixel to specific cluster depends on the distance of the pixel to the centre of the cluster in the feature domain. This method has produced better results than K and C means algorithm.

- **Proposed Methodology**
    - Resizing of all MRIs (to be analyzed) to bring them at same size using Interpolation Methods
    - Registration of MRIs using Biomarkers as Control Points
    - Comparing the size of the Target MRI (MRI of the potential Alzheimer's Patient) with MRI of a normal person of the same age
    - Identification of the Target Area (part of brain to be analyzed)
    - Identification of Black (Cavity) and White (not Black) cells
    - Calculating the pixel values of the Black and White Arrays
    - Analysis of the Target MRI using Statistical Methods
    - Report Generation